\documentclass[10pt,twocolumn,letterpaper]{article}

\usepackage{wacv}
\usepackage{times}
\usepackage{epsfig}
\usepackage{graphicx}
\usepackage{amsmath}
\usepackage{amssymb}

\usepackage{nomencl}                    
\usepackage{float}                      
\usepackage[numbers]{natbib}                     
\usepackage{graphicx}					
\usepackage{fancyhdr}                   
\usepackage{url}                        
\usepackage[inactive]{srcltx}		 	
\usepackage{relsize}                    
\usepackage{booktabs}                   
\usepackage[config, labelfont={bf}]{caption}
\usepackage{mathrsfs}                   
\usepackage[titletoc]{appendix}			
\usepackage{pdflscape}					
\usepackage{multirow}
\usepackage{dsfont}
\usepackage{amsfonts}
\usepackage{mathtools} 

\usepackage{amsthm}
\usepackage{booktabs}
\usepackage{tabularx}
\usepackage{algorithm}
\usepackage{algorithmic}
\usepackage{longtable}

\usepackage[update,prepend]{epstopdf}   
\graphicspath{figures/}
\usepackage{subcaption}

\wacvfinalcopy 

\def\wacvPaperID{136} 

\ifwacvfinal\fi
\begin{document}

\title{Discriminative Cross-View Binary Representation Learning}

\author{Liu Liu, Hairong Qi\\
Department of Electrical Engineering and Computer Science\\
University of Tennessee, Knoxville\\
{\tt\small \{lliu25, hqi\}@utk.edu}
}

\maketitle
\ifwacvfinal\thispagestyle{empty}\fi

\begin{abstract}
Learning compact representation is vital and challenging for large scale multimedia data. Cross-view/cross-modal hashing for effective binary representation learning has received significant attention with exponentially growing availability of multimedia content. Most existing cross-view hashing algorithms emphasize the similarities in individual views, which are then connected via cross-view similarities. 
In this work, we focus on the exploitation of the discriminative information from different views, and propose an end-to-end method to learn semantic-preserving and discriminative binary representation, dubbed Discriminative Cross-View Hashing (DCVH), in light of learning multitasking binary representation for various tasks including cross-view retrieval, image-to-image retrieval, and image annotation/tagging. 
The proposed DCVH has the following key components. First, it uses convolutional neural network (CNN) based nonlinear hashing functions and multilabel classification for both images and texts simultaneously. Such hashing functions achieve effective continuous relaxation during training without explicit quantization loss by using Direct Binary Embedding (DBE) layers. Second, we propose an effective view alignment via Hamming distance minimization, which is efficiently accomplished by bit-wise XOR operation. Extensive experiments on two image-text benchmark datasets demonstrate that DCVH outperforms state-of-the-art cross-view hashing algorithms as well as single-view image hashing algorithms. In addition, DCVH can provide competitive performance for image annotation/tagging. 
\end{abstract}

\section{Introduction}

Representation learning provides key insights and understanding of visual content, and thus is vital to computer vision tasks. On one hand, due to the increasing availability of image content, image hashing methods have been proposed to learn compact binary hash codes for similarity search purpose. Usually image hashing methods aim to project images onto the Hamming space where semantic similarity in the original space is well preserved. This is often realized via pairwise similarity~\cite{Liu_2016_CVPR,dhn} or triplet loss~\cite{Zhuang_2016_CVPR}. Several recent works also reveal that high-quality binary codes can be learned via classification task~\cite{cebits,Shen_2015_CVPR,dbe}, and the learned codes provide great performance for both retrieval task and classification task. On the other hand, rapidly growing social media offer massive volumes of multimedia content as well, e.g., photo posts with textual tags on Flickr, tweets with pictures, etc. It is desired to perform efficient content understanding and analytics across different media modalities. Particularly, we are interested in understanding multimedia data involving images and textual information involving tags/annotations. To this end, cross-view hashing, or cross-modal hashing has been studied, and drawing great attention~\cite{cvh,cmssh,cmfh,seph,acq,dcmh,dvsh}. 
\begin{figure}[t]
	\centering
    		  \includegraphics[width=2.8in]{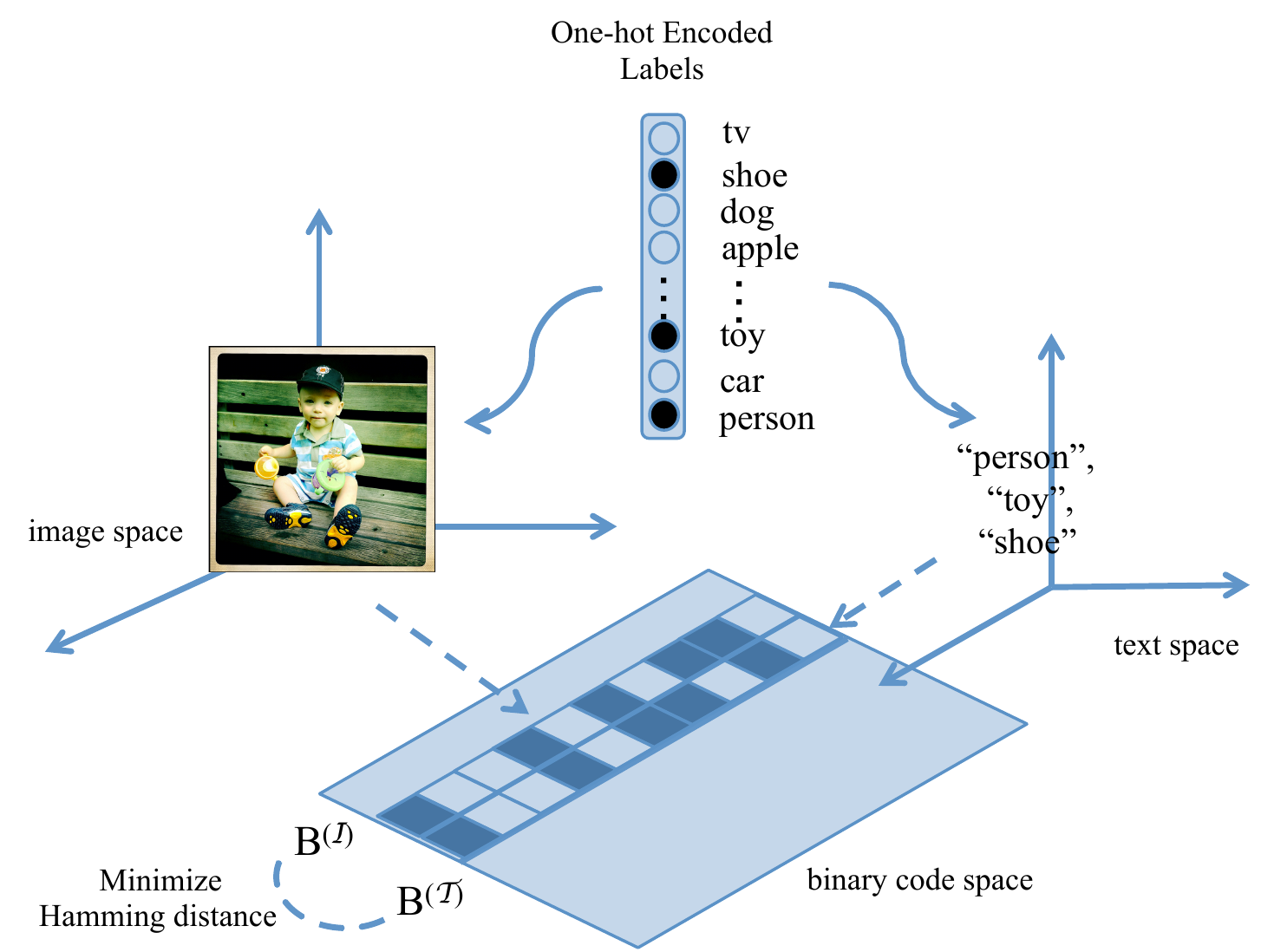}
  	\caption{Rather than using cross-view similarities, DCVH learn discriminative view-specific binary representation via multilabel classification and align them by Hamming distance minimization via bit-wise XOR.}
\label{fig:concept}
\end{figure}

Cross-view hashing studies the heterogeneous relationship of different views of data (e.g. images and the associated textual information), attempting to map them into common Hamming space. This enables both inter- and intra-indexing of data from different views. Previous methods mainly rely on a similarity matrix~\cite{dcmh,THN,cvh,cmssh} or provided affinities~\cite{seph,acq} to delineate the cross-view relationship. However, representing such similarity is quite resource-intensive as its size tends to grow quadratically with the size of the training set; furthermore, the discriminative information that is useful to other tasks such as single-view image hashing and annotation is not well preserved by the similarities. 

To address these problems, we propose Discriminative Cross-View Hashing (DCVH), an end-to-end approach, to learn semantic-preserving and {\it discriminative} binary representation. As illustrated in Fig.~\ref{fig:concept}, DCVH adopts multilabel classification directly to learn discriminative view-specific binary representation. Explicitly, a deep convolutional neural network~\cite{He_2016_CVPR} projects images into lower-dimensional latent feature space; for texts, DCVH uses pretrained GloVe~\cite{pennington2014glove} vectors to represent words in the texts. Then vector representation for each textual instance is formed by concatenating GloVe vectors. They are fed to a text-CNN~\cite{text_cnn}, mapping text vectors into a common latent feature space of images. Meanwhile, a Direct Binary Embedding (DBE) layer~\cite{dbe} is employed in both deep CNN and text-CNN, enabling the learning of binary representations without explicit quantization loss. Finally, DCVH aligns different views by minimizing the Hamming distance between the view-specific binary representations directly. This is efficiently accomplished by bit-wise XOR operation. Extensive experiments are conducted to validate the proposed DCVH, and the results demonstrate that DCVH improves cross-view retrieval and single-view image retrieval tasks over state-of-the-art by a large margin. Meanwhile DCVH can also provide competitive performance on image annotation/tagging task, suggesting that DCVH learns multitasking binary representations. The contributions of this work can be outlined as follows.
\begin{enumerate}
\itemsep0em 
\item We propose to use CNN based nonlinear projections for both image view and text view. By using the Direct Binary Embedding (DBE) layers in both projections, no quantization loss is needed to achieve effective continuous relaxation during training.
\item We achieve effective view alignment, which directly minimizes the Hamming distance between the view-specific binary representation via bit-wise XOR operation, thanks to the inclusion of the DBE layer.
\item The end-to-end DCVH learns semantic-preserving and discriminative binary representation for images and texts simultaneously, leading to multitasking binary representation that can be used as high-quality hash code both for retrieval purpose, and compact image features for classification/annotation purpose.
\end{enumerate}

The remainder of this paper is organized as follows. Section~\ref{sec:related} discusses related work and their impact on the proposed method. Section~\ref{sec:dcvh} presents DCVH in details, including view-specific learning and view alignment. Section~\ref{sec:exp} reports the results of extensive experiments to validate DCVH on various tasks, i.e., cross-view retrieval, single-view image retrieval, and image annotation/tagging. Finally Section~\ref{sec:con} summarizes this paper and presents concluding remarks.


\section{Related Works}\label{sec:related}
\begin{figure*}
	\centering
    		  \includegraphics[width=6.2in]{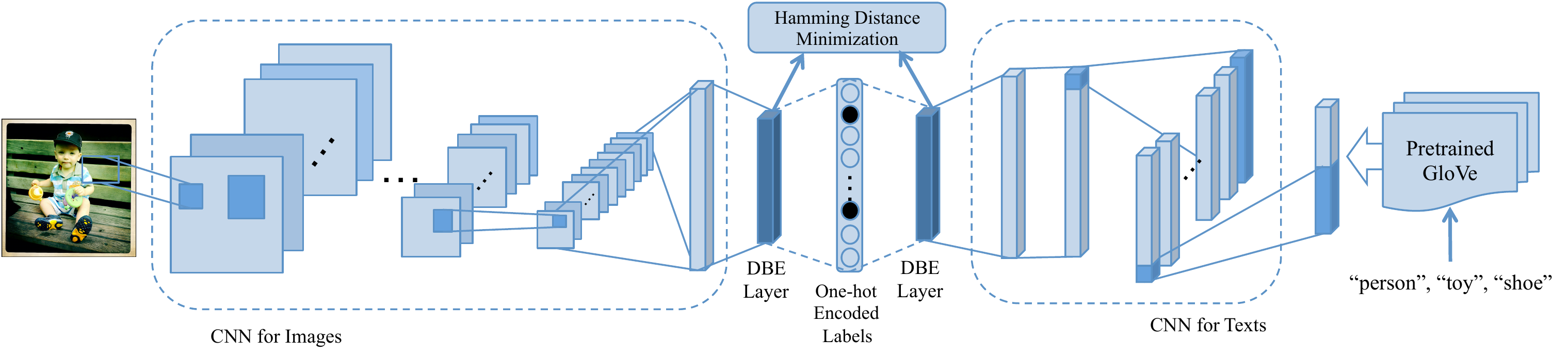}
  	\caption{The overview architecture of the proposed Discriminative Cross-View Hashing (DCVH). Given multimedia data (images-texts), DCVH uses convolutional neural network (CNN)~\cite{He_2016_CVPR} to project images into binary representation; meanwhile DCVH uses pretrained GloVe~\cite{pennington2014glove} vectors to obtain text vector representation. Then text vectors are fed into a text-CNN~\cite{text_cnn} to generate binary representation. Unlike most methods that uses cross-view similarities, DCVH uses multilabel classification to embed raw images and texts into common binary feature space. Hammning distance minimization is adopted for view alignment purpose}
\label{fig:arch}
\vspace{-2mm}
\end{figure*}

Our work is closely related to cross-view hashing or cross-modal hashing. Majority of cross-view hashing methods are based on hand-crafted features, and cannot provide end-to-end solutions. For instance, CCA~\cite{Hardoon:2004:CCA:1119696.1119703} minimizes the distance between paired modalities in a latent space on the condition that their norm is equal to one; CVH~\cite{cvh} is a cross-modal extension of Spectral Hashing~\cite{spectralhashing}. CMSSH~\cite{spectralhashing} maps multi-view data into common Hamming space by solving binary classification problem. CMFH~\cite{cmfh} uses matrix factorization to solve the cross-modal hashing problem. ACQ\cite{acq} minimizes quantization error for different modalities alternatively while preserving data similarities. SePH~\cite{seph} transforms affinities into global probabilities and learns view-invariant codes via KL-divergence. Recently several deep end-to-end cross-view hashing approaches have been proposed. For example, DVSH~\cite{dvsh} employs CNN and LSTM for both images and sentences modality, respectively, while preserving the pairwise similarity via cosine max-margin loss. THN\cite{THN} proposes a hybrid deep architecture and use auxiliary datasets to learn cross-modal correlation for heterogeneous modalities using pairwise cross-entropy loss. DCMH~\cite{dcmh} uses CNN and neural network (NN) for embedding images and texts separately and connect them with cross-view similarities. Similar to the hand-crafted feature based methods, these deep approaches also focus on cross-view similarities without considering view-specific discriminative information. This is different from our perspective on cross-view retrieval, which is presented in DCVH by using classification explicitly in separate views, and aligning the view by minimizing Hamming distance between the binary representations. This leads semantic-preserving and discriminative binary representation not only useful for cross-view retrieval, but also capable of single-view similarity search and image annotation/tagging tasks.

Our work is also related to image hashing via classification and multilabel image classification. Previous works such as SDH~\cite{Shen_2015_CVPR} and DBE~\cite{dbe} provide evidence that binary codes learned through classification tasks serve as strong hash code for retrieval tasks and image features for classification purpose, thanks to the discriminative information obtained via classification. This inspires us to adopt multilabel classification to learn discriminative binary representation for both images and texts, thus potentially competent for various tasks. Meanwhile, multilabel classification has attracted much attention. WARP~\cite{warp} uses a ranking-based approach together with a CNN for image annotation. CNN-RNN~\cite{Wang_2016_CVPR} and DBE~\cite{dbe} suggest that binary cross entropy provides strong performance for classification task despite its simplicity. Consequently, we adopt binary cross entropy as the loss function.

Comparing to images, representation learning and classification for texts are solved differently. Since most natural language processing (NLP) problems deal with sequential data, e.g., sentences and documents, recurrent neural networks (RNNs) such as Long Short-Term Memory (LSTM) are used~\cite{sentence_lstm} due to their capability of capturing long-term dynamics. Meanwhile, most cross-view hashing methods adopt bag-of-word (BoW) to represent textual information~\cite{seph,dcmh,cvh,THN}. Although simple, the rich semantic information of texts might not be well preserved. Alternatively, ACQ~\cite{acq} represents texts with mean vectors of word vectors from word2vec features with linear transformation, leading to similar problem as well. Recent works suggest that CNNs are effective to solve NLP problems as well. For instance, THC~\cite{text_cnn} adopt one-dimensional CNN to learn binary representation of texts by using word features and position features.  Meanwhile, most real-world images available through social media are associated with tags or annotations, where the sequential structure is not a strong as that of sentences. As suggested by fastText~\cite{joulin2016bag}, texts can be conveniently modeled by a linear classifier with a hidden variable provided a lookup table for words. This inspires us to adopt a word embedding lookup table to embed words into vectors. 1-D CNN is then employed to learn binary representation, similar to that for images.

\section{Proposed Algorithm}\label{sec:dcvh}

The proposed Discriminative Cross-View Hashing (DCVH) is presented in this section. It consists of three components: a deep structure that maps images into low-dimensional Hamming space; a lookup table for text vector representation followed by a text-CNN to embed textual information into common Hamming space; and a view alignment that minimizes the Hamming distance between corresponding image and text pair. The overview architecture of DCVH is illustrated in Fig.~\ref{fig:arch}.

In cross-view image hashing, a training set $\mathcal{S} = \{s_i\}_{i=1}^N$ consisting $N$ instances is provided, where each instance in $\mathcal{S}=(\mathcal{I},\mathcal{T})$ has two corresponding views: image-view $\mathcal{I}=\{I_i\}_{i=1}^N$ and text-view $\mathcal{T}=\{T_i\}_{i=1}^N$. We aim to generate semantic-preserving and discriminative binary representation $\mathbf{B}^{(\mathcal{S})}\in \{0,1\}^{N\times D}$ for ${(\mathcal{S})}$ from the two views by the following hashing functions
\begin{align}\label{eq:hash}
&\mathbf{B}^{(\mathcal{S})} = thresold(F^{(\mathcal{S})}(\mathcal{S};\Omega^{(\mathcal{S})}),0.5),\; \mathcal{S} = \mathcal{I}\ \text{or}\ \mathcal{T}\\
&\text{s.t.}\quad threshold(v,t) =
	\begin{cases}
	1,&\quad v\geq t\\
	0,&\quad \text{otherwise}
	\end{cases}\nonumber
\end{align}
where $F^{(\mathcal{S})}(\cdot;\Omega^{(\mathcal{S})})$ is nonlinear projections parameterized by $\Omega^{(\mathcal{S})}$ for image-view $\mathcal{S} = \mathcal{I}$ and text-view $\mathcal{S} = \mathcal{T}$.

Discriminative information that could be useful for other tasks might not be well preserved via similarity-based supervision. Therefore, we adopt the paradigm of classification-based binary code learning~\cite{Shen_2015_CVPR,dbe}, and use textual information to obtain labels to classify both images and texts. One direct approach of generating the labels is to encode the text into one-hot labels according to whether a tag or annotation for an instance appears or not. And the label information is denoted as $\mathcal{Y} = \{y_i\}_{i=1}^N$.

\subsection{CNN-based Hash Function for Images and Texts}
\begin{figure}[h]
	\centering
    		  \includegraphics[width=3.in]{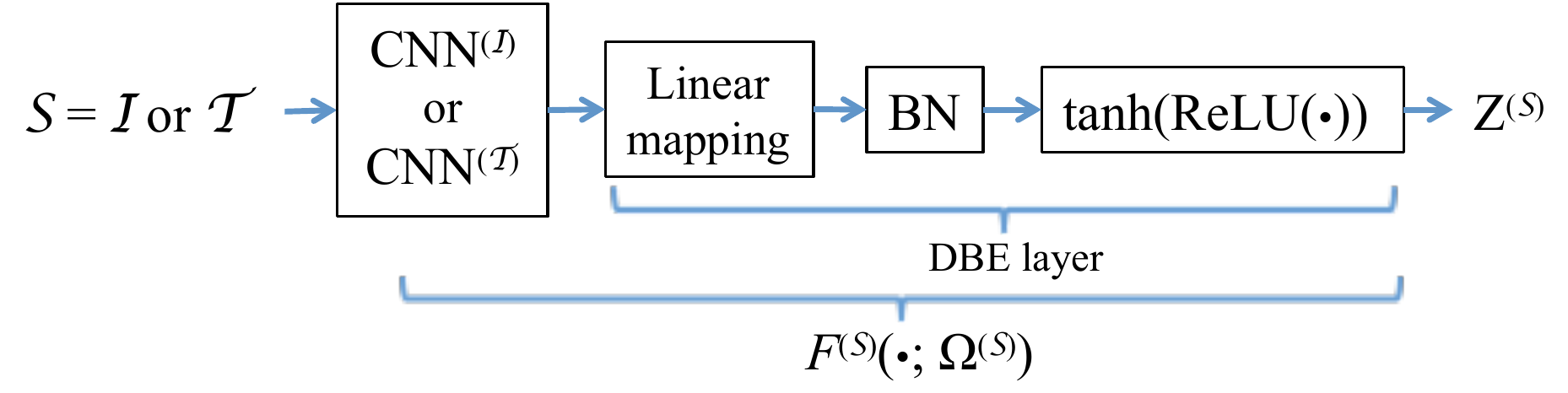}
  	\caption{Direct Binary Embedding (DBE) layer architecture and how it is concatenated to a convolutional neural network (CNN) for images or texts. $\text{CNN}^{(\mathcal{I})}$ and $\text{CNN}^{(\mathcal{T})}$ are the corresponding CNN for images and text. DBE layer uses a linear transform and batch normalization (BN) with compound nonlinearity: rectified linear unit (ReLU) and tanh, to transform CNN activation to binary-like latent feature space $Z^{(\mathcal{S})}$}
\label{fig:dbe}
\end{figure}

Proper choices of nonlinear projection $F^{(\mathcal{I})}$ and $F^{(\mathcal{T})}$ would facilitate the learning of high-quality binary representation significantly. In this work, we choose CNN based projections for both image-view and text-view. Specifically, a deep CNN (e.g., ResNet~\cite{He_2016_CVPR}) concatenated with a Direct Binary Embedding (DBE)~\cite{dbe} layer for images; and a text-CNN~\cite{text_cnn} concatenated with a DBE layer for textual information are used. The DBE layer, as illustrated in Fig.~\ref{fig:dbe}, learns a continuous binary-like representation $\mathbf{Z}^{(\mathcal{S})}$ that approximates the discrete 0-1 binary code  well, i.e., $\mathbf{Z}^{(\mathcal{S})}\approx \mathbf{B}^{(\mathcal{S})}$. This effectively eliminates the need of quantization loss, originally commonly used by hashing methods~\cite{Shen_2015_CVPR,acq,dcmh,dhn}.
\vspace{-2mm}

For texts, we employ a 2-conv layer text-CNN~\cite{text_cnn} with 1-D convolution, as demonstrated in Fig.~\ref{fig:arch}. Given vector embedding for textual information (e.g., concatenated GloVe vectors), text-CNN uses kernels with the same size as that of a GloVe vector on the first layer, and kernels with the same size as the output of the first conv layer output. Since we do not consider the sequential structure among texts, the stride size is also the same as the GloVe vector. There are 1,000 kernels on both convolutional layers and the second convolutional layer outputs 1-D vector directly. Then another fully-connected layer and a DBE-layer follow to embed the texts into a common latent space where image DBE features reside.

\subsection{View-Specific Binary Representation Learning}\label{sec:classification}
Multilabel classification is used to learn binary representation for both images and texts. We choose binary sigmoid cross entropy as the loss function for both views. Linear classifiers $\mathbf{W}^{(\mathcal{S})}$, $\mathcal{S}$ = $\mathcal{I}$ or $\mathcal{T}$, are used to classify the binary representation $\mathbf{B}^{(\mathcal{S})}$. Given the labels $\mathcal{Y} = \{y_i\}_{i=1}^N$, we have the following view-specific optimization problem:
\begin{align}\label{eq:binaryxen}
\min_{\mathbf{W}^{(\mathcal{S})},\Omega^{(\mathcal{S})}}& \mathit{L}^{(\mathcal{S})}(\mathbf{W}^{(\mathcal{S})},F^{(\mathcal{S})}(\mathcal{S};\Omega^{(\mathcal{S})}))=\\ \nonumber
 & - \frac{1}{N} \sum_{i=1}^N\sum_{p=1}^C \left[\mathds{1}(y_i)\log\frac{1}{1+e^{\mathbf{w}_p^{(\mathcal{S})\top} \mathbf{b}_i^{(\mathcal{S})}}} \right.\\ \nonumber
 &\left. + (1-\mathds{1}(y_i))\log\frac{e^{\mathbf{w}_p^{(\mathcal{S})\top} \mathbf{b}_i^{(\mathcal{S})}}}{1+e^{\mathbf{w}_p^{(\mathcal{S})\top} \mathbf{b}_i^{(\mathcal{S})}}} \right],\ \mathcal{S} = \mathcal{I}\ \text{or}\ \mathcal{T}\\
 \text{s.t.}\quad & \mathbf{b}_i^{(\mathcal{S})} = thresold(F^{(\mathcal{S})}(s_i;\Omega^{(\mathcal{S})}),0.5)\nonumber
\end{align}
where $C$ is the number of categories; $\mathbf{W}^{(\mathcal{S})}=[\mathbf{w}_1^{(\mathcal{S})}, \ldots, \mathbf{w}_C^{(\mathcal{S})}]$ and $\mathbf{w}_k^{(\mathcal{S})},\ k=1,\ldots,C$ is the weight of the classifier for category $k$; $\mathds{1}(y_i)$ an indicator function representing the probability distribution for label $y_i$.

Direct optimizing Eq.~\ref{eq:binaryxen} is difficult due to the discrete characteristics of $threshold(\cdot,\cdot)$. Meanwhile, since DBE layer approximates binary code well, Eq.~\ref{eq:binaryxen} can be relaxed to the following form without quantization loss:
\begin{align}\label{eq:xenrelaxed}
&\min_{\mathbf{W}^{(\mathcal{S})},\Omega^{(\mathcal{S})}} \mathit{L}^{(\mathcal{S})}(\mathbf{W}^{(\mathcal{S})},F^{(\mathcal{S})}(\mathcal{S};\Omega^{(\mathcal{S})}))\approx\\ \nonumber
& - \frac{1}{N} \sum_{i=1}^N\sum_{p=1}^C \left[\mathds{1}(y_i)\log\frac{1}{1+e^{\mathbf{w}_p^{(\mathcal{S})\top} F^{(\mathcal{S})}(s_i;\Omega^{(\mathcal{S})})}} \right.\\ \nonumber
&\left. + (1-\mathds{1}(y_i))\log\frac{e^{\mathbf{w}_p^{(\mathcal{S})\top} F^{(\mathcal{S})}(s_i;\Omega^{(\mathcal{S})})}}{1+e^{\mathbf{w}_p^{(\mathcal{S})\top} F^{(\mathcal{S})}(s_i;\Omega^{(\mathcal{S})})}} \right],\ \mathcal{S} = \mathcal{I}\ \text{or}\ \mathcal{T}\nonumber
\end{align}
Eq.~\ref{eq:xenrelaxed} suggests that the learning of the nonlinear projections and the classifiers are optimized in the end-to-end fashion.

\subsection{View Alignment}
For the purpose of effective cross-view indexing, it is necessary to align the learned binary representations from the two views. In order to do so, we propose to directly minimize the distance between the learned binary representations. As a common distance metric for binary codes, Hamming distance can be conveniently expressed as \textit{XOR} operation between two codes. Therefore, we attempt to minimize the Hamming distance between two corresponding binary representation $\mathbf{B}^{\mathcal{I}}$ and $\mathbf{B}^{\mathcal{T}}$ in order to achieve effective view alignment:
\begin{align}\label{eq:align}
\min_{\Omega^{(\mathcal{I})},\Omega^{(\mathcal{T})}}& \mathit{J}_{\mathcal{I},\mathcal{T}}(F^{(\mathcal{I})}(\mathcal{I};\Omega^{(\mathcal{I})}),F^{(\mathcal{T})}(\mathcal{T};\Omega^{(\mathcal{T})}))\\ \nonumber
=& \frac{1}{ND}\sum\mathbf{B}^{\mathcal{I}}\oplus \mathbf{B}^{\mathcal{T}}\\ \nonumber
=&\frac{1}{ND}\sum  \mathbf{B}^{\mathcal{I}}\odot\overline{\mathbf{B}}^{\mathcal{T}}+\overline{\mathbf{B}}^{\mathcal{I}}\odot\mathbf{B}^{\mathcal{T}}\\
\text{s.t.}\;& \mathbf{B}^{(\mathcal{S})} = thresold(F^{(\mathcal{S})}(\mathcal{S};\Omega^{(\mathcal{S})}),0.5),\ \mathcal{S} = \mathcal{I}\ \text{or}\ \mathcal{T}\nonumber
\end{align}
where $\oplus$ is bit-wise XOR; $\odot$ is Hadamart multiplication or element-wise product. Since $\mathbf{B}^{(\mathcal{I})}$ and $\mathbf{B}^{(\mathcal{T})}$ are both binary, $\odot$ is equivalent to bit-wise AND.
Similar to Section~\ref{sec:classification}, Eq.~\ref{eq:align} can be relaxed as:
\begin{align}
\min_{\Omega^{(\mathcal{I})},\Omega^{(\mathcal{T})}}& \mathit{J}_{\mathcal{I},\mathcal{T}}(F^{(\mathcal{I})}(;\Omega^{(\mathcal{I})}),F^{(\mathcal{T})}(;\Omega^{(\mathcal{T})}))\approx\\ \nonumber
 & \frac{1}{ND} \sum_{i=1}^N \Big\{F^{(\mathcal{I})}(I_i;\Omega^{(\mathcal{I})})(\mathbf{1}-F^{(\mathcal{T})}(T_i;\Omega^{(\mathcal{T})}))^\top\\ \nonumber
&+(\mathbf{1}-F^{(\mathcal{I})}(I_i;\Omega^{(\mathcal{I})}))F^{(\mathcal{T})}(T_i;\Omega^{(\mathcal{T})})^\top\Big\}
\end{align}

\subsection{Total Formulation and Algorithm }
DCVH learns cross-view binary representations by combining the view-specific learning and view alignment. Then the final formulation is
\begin{align}\label{eq:total}
\min_{\mathclap{\substack{\mathbf{W}^{(\mathcal{I})},\mathbf{W}^{(\mathcal{T})},\\ \Omega^{(\mathcal{I})},\Omega^{(\mathcal{T})}}}} \quad &(1-\lambda)\left(\mathit{L}^{(\mathcal{I})}+\mathit{L}^{(\mathcal{T})}\right) + \lambda \mathit{J}_{\mathcal{I},\mathcal{T}}, \\
\text{s.t.}\quad &\lambda\in (0,1]\nonumber
\end{align}
where $\lambda$ is a hyperparameter introduced to control the degree of the view alignment. Intuitively, the higher $\lambda$, the more matching two views will be, but the learned binary representation will be less discriminative; and vice versa. The detailed discussion on $\lambda$ is included in Section~\ref{sec:hyper}.

The training of DCVH is the same as that of regular CNN, where stochastic gradient descent (SGD) is used to iterate through mini-batches of training data. In order to accelerate the training process, $F^{(\mathcal{I})}(\cdot;\Omega^{(\mathcal{I})})$ and $F^{(\mathcal{T})}(\cdot;\Omega^{(\mathcal{T})})$ are pretrained separately before the optimization of Eq.~\ref{eq:total}. Formally, DCVH is presented in Algorithm~\ref{alg:dcvh}.

\begin{algorithm}
\caption{The Training of DCVH}\label{alg:dcvh}
\begin{algorithmic}[0]
\STATE \textbf{Input}: The training set $\mathcal{S}$ consisting of corresponding images $\mathcal{I}$, texts $\mathcal{T}$ and the one-hot labels $\mathcal{Y}$ obtained from texts.
\STATE \textbf{Output}: Parameters of nonlinear projections $F^{(\mathcal{I})}(\cdot;\Omega^{(\mathcal{I})})$, $F^{(\mathcal{T})}(\cdot;\Omega^{(\mathcal{T})})$, and linear classifier $\mathbf{W}^{(\mathcal{I})},\mathbf{W}^{(\mathcal{T})}$.
\STATE 
\STATE \textbf{Initialization}: Initialize vector embedding of texts with pretrained GloVe embedding. Initialize learning rate with $\mu$, mini-batch size with 64, and maximum iterations with ITER.
\STATE \textbf{Pretraining}: Pretrain the nonlinear projections $F^{(\mathcal{I})}(\cdot;\Omega^{(\mathcal{I})})$ with $\mathbf{W}^{(\mathcal{I})}$, and $F^{(\mathcal{T})}(\cdot;\Omega^{(\mathcal{T})})$ with $\mathbf{W}^{(\mathcal{T})}$ separately according to Eq.~\ref{eq:xenrelaxed}.
\FOR {ITER}
\STATE Update $\Omega^{(\mathcal{I})}$ by\\
				\begin{center}$\Omega^{(\mathcal{I})} \leftarrow \Omega^{(\mathcal{I})}-\mu \frac{\partial}{\partial \Omega^{(\mathcal{I})}}\left((1-\lambda)\mathit{L}^{(\mathcal{I})}+ \lambda \mathit{J}\right)$\end{center}
				
\STATE Update $\Omega^{(\mathcal{T})}$ by\\
				\begin{center}$\Omega^{(\mathcal{T})} \leftarrow \Omega^{(\mathcal{T})}-\mu \frac{\partial}{\partial \Omega^{(\mathcal{T})}}\left((1-\lambda)\mathit{L}^{(\mathcal{T})}+ \lambda \mathit{J}\right)$\end{center}
\STATE Update $\mathbf{W}^{(\mathcal{I})}$ by\\
				\begin{center}$\mathbf{W}^{(\mathcal{I})} \leftarrow \mathbf{W}^{(\mathcal{I})}-\mu(1-\lambda) \frac{\partial}{\partial \mathbf{W}^{(\mathcal{I})}}\mathit{L}^{(\mathcal{I})}$\end{center}
\STATE Update $\mathbf{W}^{(\mathcal{T})}$ by\\
				\begin{center}$\mathbf{W}^{(\mathcal{T})} \leftarrow \mathbf{W}^{(\mathcal{T})}-\mu(1-\lambda) \frac{\partial}{\partial \mathbf{W}^{(\mathcal{T})}}\mathit{L}^{(\mathcal{T})}$\end{center}
\ENDFOR
\end{algorithmic}
\end{algorithm}

For a test sample that is not in the training set, the binary representation can be obtained via Eq.~\ref{eq:hash}. And retrieving from database can be efficiently performed by ranking the retrieved results according the Hamming distance; or using pre-defined Hamming radius to perform hash lookup. For tagging/annotation purpose, the predicted tags or annotations for the test sample image can be obtained from the top-$k$ predictions by using the classifier $\mathbf{W}^{(\mathcal{I})}$, where $k$ is the number of tags or annotations. This overlaps with the task of retrieving textual information given test sample images, but it is more accurate since usually a retrieval is considered successful as long as one tag/annotation matches, where multilabel classification requires top-$k$ predication matches simultaneously.

\subsection{Extensibility}
Although DCVH is proposed in the discussion of cross-view data, it is easily extended to the circumstances where three or more views of data is available. If there are $m$ views of data, i.e., $\mathcal{S} = (S_1, S_2, \ldots, S_m)$. A direct way to extend Eq.~\ref{eq:total} is by considering mutual Hamming distance minimization:
\begin{align}\label{eq:total}
\min_{\mathclap{\mathbf{W}^{(S_i)},\Omega^{(S_i)}, \forall i}}\quad &(1-\lambda)\sum_{i=1}^m\mathit{L}^{(S_i)} + \lambda \sum_{i,j}\mathit{J}_{i,j},\\
\text{s.t.}\quad &\lambda\in (0,1]\nonumber
\end{align}
where $\mathit{J}_{i,j}$ is the Hamming distance between the binary representation of $i$th and $j$th view.

\section{Experiments}\label{sec:exp}
\subsection{Datasets}
We evaluate the proposed DCVH on two image-text benchmark datasets: MS COCO~\cite{Lin2014} and MIRFLICKR~\cite{huiskes08}.

\textbf{MS COCO} is a dataset for image recognition, segmentation and captioning. It contains a training set of 83K images and a validation set of 40K images. There are totally 80 categories of annotations with several annotations per image. Since images in COCO are color images with various sizes, we resize them to $224\times 224$. The textual representation for DCVH are obtained by concatenating pretrained GloVe~\cite{pennington2014glove} vectors. And all the text vectors are zero-padded to the same size, resulting in a 6000-D vector per image. Text vectors for the compared algorithms are obtained from word2vec~\cite{word2vec} vectors, as suggested by ACQ~\cite{acq}. For the hand-crafted feature based algorithms, images are represented by AlexNet~\cite{DBLP:conf/nips/KrizhevskySH12} activation features; raw images are used for end-to-end algorithms.

\textbf{MIRFLICKR} contains 25K color images originally collected from Flickr, and each image is associated with several textual tags. All the images are also resized to $224\times 224$. Following the settings of DCMH~\cite{dcmh}, we remove the images with noisy textual information and without valid textual information. Similarly, the textual information for each image is represented by a 4200-D vector from pretrained GloVe embeddings for the proposed DCVH. For the compared methods, the texts for the comparing algorithms is represented as BoW vectors as suggested by them.

\subsection{Experimental Settings and Protocols}
Several state-of-the-art methods for cross-view hashing, supervised image hashing and image annotation are adopted as baselines to compare with DCVH. Since real-world images usually contain multiple tags or visual concepts, we consider a retrieval is successful when the retrieved item shares at least one concept with the query. We set hyperparameter $\lambda=0.2$ of DCVH throughout the experiments. Learning rate $\mu$ is set to $0.0002$ for pretraining, and $0.0001$ with exponential decay for further training. We pretrain ResNet-50~\footnote{We use TensorFlow Slim library ResNet-50 model. We freeze the batch normalization layers of ResNet-50 as recommended during fine-tuning for easier convergence, and find that such ResNet provide comparable performance as AlexNet with batch normalization.} for images 10,000 iterations and text-CNN for texts 2,000 iterations; and further train them together with view alignment 10,000 iterations. DCVH is implemented in TensorFlow~\cite{45166}.

\begin{table*}
\centering
\scalebox{1.0}{
\begin{tabular}{ c | l | c c c | c c c  }
	\toprule
	\multirow{2}{*}{Task}	& 	\multirow{2}{*}{Method}	& \multicolumn{3}{c|}{MS COCO} & \multicolumn{3}{c}{MIRFLICKR}  \\
	\cline{3-8}
							&							& 16 bits & 32 bits & 64 bits & 16 bits & 32 bits & 64 bits\\
	\hline
	\multirow{8}{*}{Image Query, Text Database}& CVH~\cite{cvh}    		& 0.484 & 0.471 & 0.435 & 0.607 & 0.618 & 0.616\\	
												& CMSSH~\cite{cmssh}		& 0.472 & 0.467 & 0.453 & 0.573 & 0.574 & 0.571\\ 
												& CMFH~\cite{cmfh} 		& 0.486 & 0.517 & 0.545 & 0.586 & 0.584 & 0.584\\  
												& SePH~\cite{seph}		& 0.543	& 0.551 & 0.557 & 0.672 & 0.677 & 0.679\\
												& ACQ~\cite{acq}			& 0.531 & 0.544 & 0.555 &  -	    &   -   &  -  \\
												& DCMH~\cite{dcmh}		&    -  &   -   &   -   & \bf{0.741} & \underline{0.747} & \underline{0.749} \\
												\cline{2-8}
												& DCVH-THN				& 0.601 & 0.618 & 0.623 & 0.681 & 0.692 & 0.706\\
												& DCVH-CVH				& \underline{0.703}	& \underline{0.721}	& \underline{0.728}	& 0.710 & 0.723 & 0.728\\
												& DCVH		& {\bf 0.710} & {\bf 0.728} & {\bf 0.733} &  \underline{0.735} & {\bf 0.755} & {\bf 0.769} \\
	\hline
	\hline
	\multirow{8}{*}{Text Query, Image Database}& CVH~\cite{cvh}    		& 0.480 & 0.467 & 0.432 & 0.603 & 0.604 & 0.602\\	
												& CMSSH~\cite{cmssh}		& 0.465 & 0.454 & 0.446 & 0.572 & 0.573 & 0.570\\ 
												& CMFH~\cite{cmfh} 		& 0.486 & 0.517 & 0.545 & 0.586 & 0.584 & 0.584\\  
												& SePH~\cite{seph}		& 0.549	& 0.557 & 0.562 & 0.720 & 0.727 & 0.731\\
												& ACQ~\cite{acq}			& 0.521 & 0.546 & 0.562 &  -	    &   -   &  -  \\
												& DCMH~\cite{dcmh}		&    -  &   -   &   -   & \underline{0.783} & \underline{0.790} & 0.793 \\
												\cline{2-8}
												& DCVH-THN				& 0.619 & 0.631 & 0.645 & 0.719 & 0.734 & 0.743\\
												& DCVH-CVH				& \underline{0.728} & \underline{0.749}	& \underline{0.753} & 0.778 & 0.785 & \underline{0.796}\\
												& DCVH 			& {\bf 0.739} & {\bf 0.757} & {\bf 0.763} & {\bf 0.812} &  {\bf 0.828} & {\bf 0.840} \\
	\bottomrule
\end{tabular}
}
\caption{mAP values for the proposed DCVH and compared baselines on cross-view retrieval task with all the benchmark datasets. Results are provided for different code lengths.}
\label{tab:map_results}
\end{table*}

\begin{figure*}
    \centering
    \begin{subfigure}[h]{0.23\textwidth}
        \centering
        \includegraphics[height=1.15in]{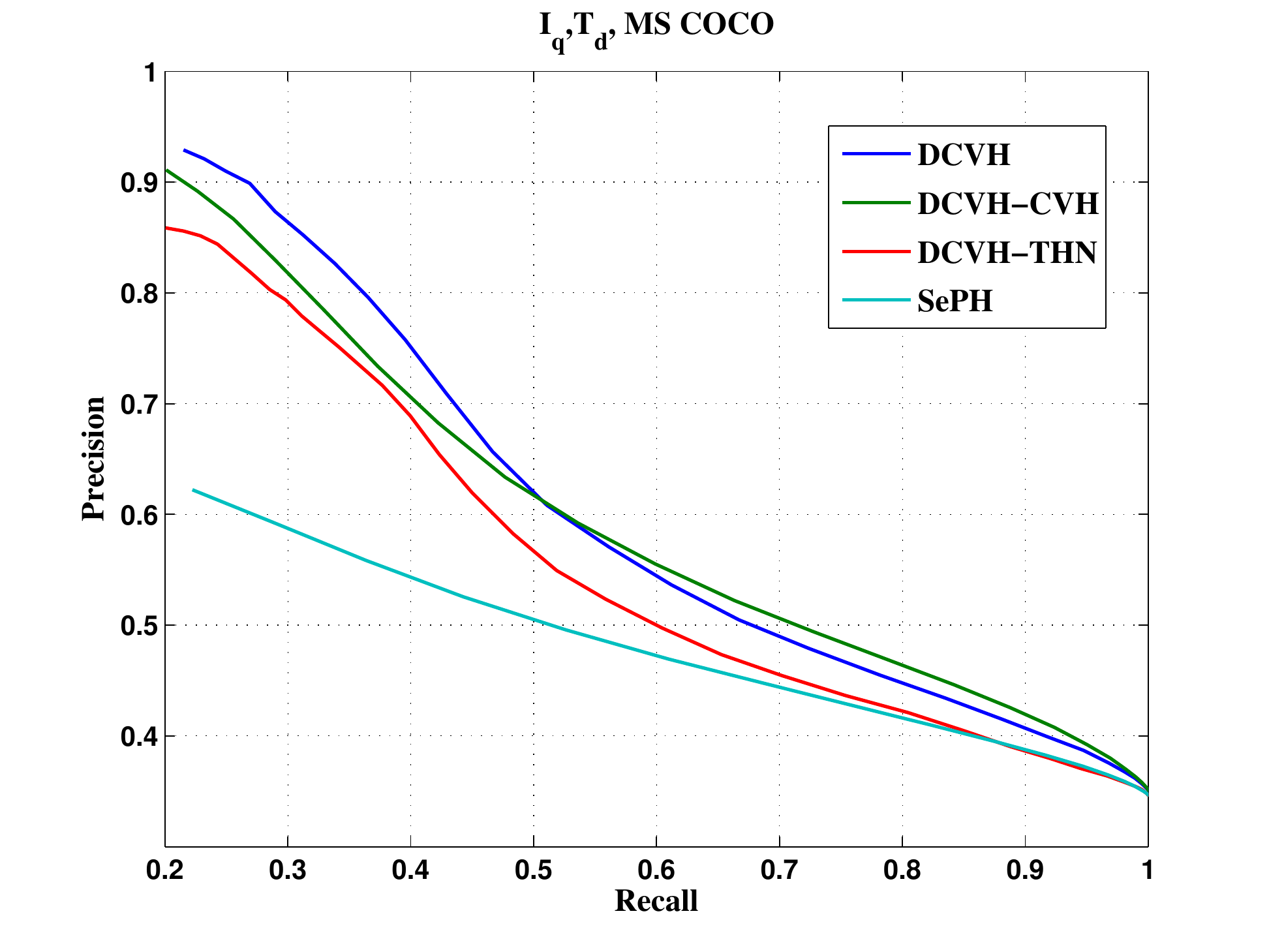}
        \caption{I-Q, T-D, MS COCO}
       	\label{fig:iqtd_coco}
    \end{subfigure}%
    \begin{subfigure}[h]{0.23\textwidth}
        \centering
        \includegraphics[height=1.15in]{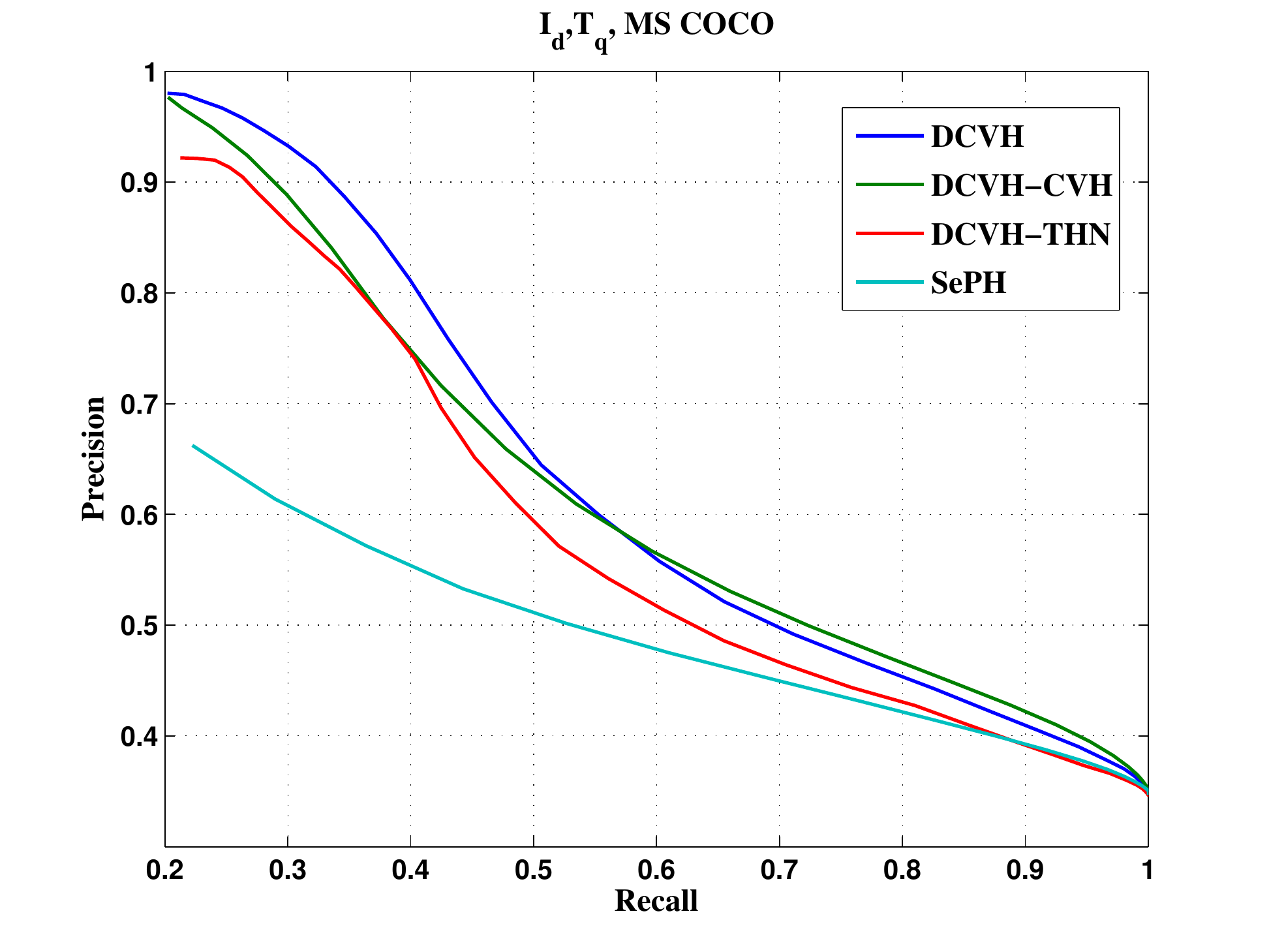}
        \caption{I-D, T-Q, MS COCO}
        \label{fig:idtq_coco}
    \end{subfigure}
    \begin{subfigure}[h]{0.23\textwidth}
        \centering
        \includegraphics[height=1.15in]{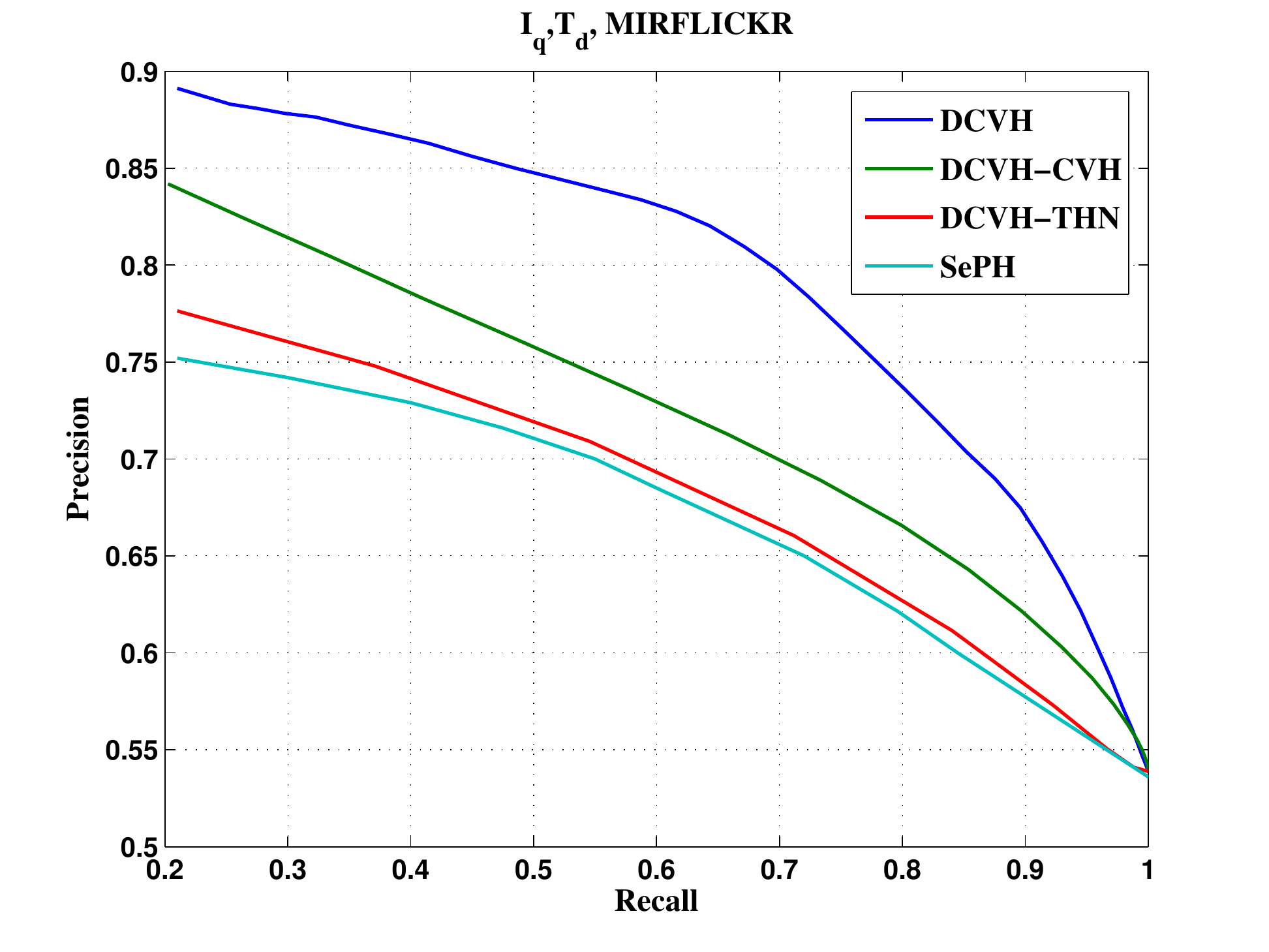}
        \caption{I-Q, T-D, MIRFLICKR}
       	\label{fig:iqtd_coco}
    \end{subfigure}%
    \begin{subfigure}[h]{0.23\textwidth}
        \centering
        \includegraphics[height=1.15in]{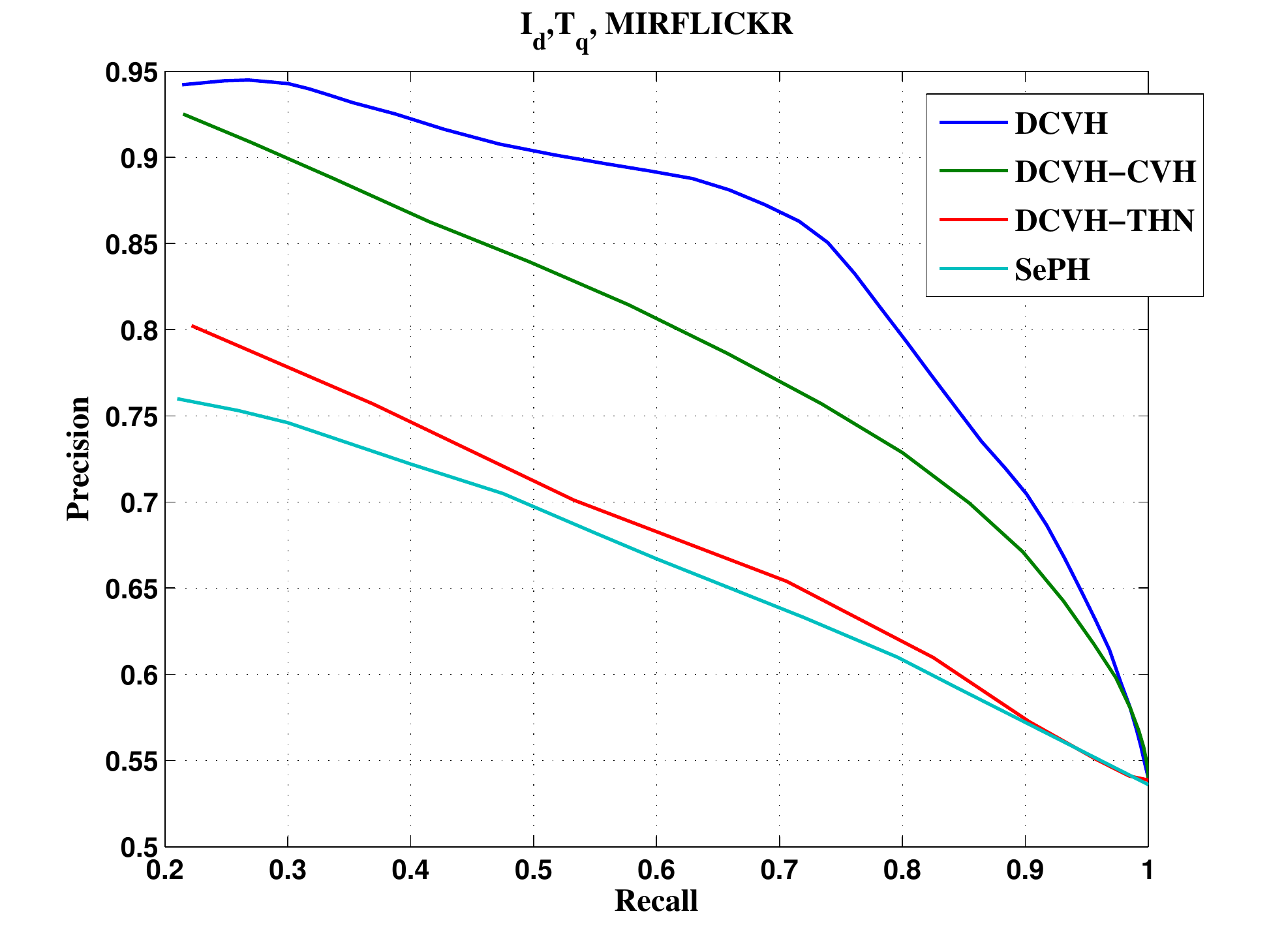}
        \caption{I-D, T-Q, MIRFLICKR}
        \label{fig:idtq_coco}
    \end{subfigure}
    \caption{Comparison of precision-recall curve with code length of 64 bits on tasks: image query with text dataset (I-D, T-D) and text query with image dataset (I-D, T-D), on MS COCO ((a), (b)), and MIFLICKR ((c), (d)).}
\label{fig:prcurve}
\end{figure*}

For the MS COCO dataset, we choose to compare with several state-of-the-art hand-crafted feature based algorithms including CVH~\cite{cvh}, CMSSH~\cite{cmssh}, CMFH~\cite{cmfh}, SePH~\cite{seph}, ACQ~\cite{acq}. We also compare with end-to-end approach. Since the code for most deep algorithms is not available, we adopt the pairwise cross-entropy loss provided in THN~\cite{THN}, and weighted average Hamming distance between different views in CVH~\cite{cvh} as the view alignment, and propose two variations of DCVH, denoted as DCVH-THN and DCVH-CVH. 1,000 samples are randomly selected to form the query set and the rest is treated as the database for retrieval purpose and training. Note that DCVH and its variations are trained on the provided training set only. In addition, we compare with several state-of-the-art supervised image hashing algorithms for the purpose of showing DCVH is able to improve single-view image hashing performance as well. 5,000 images are randomly selected as the query set. We choose to compare with HashNet~\cite{hashnet}, DBE~\cite{dbe}, DHN~\cite{dhn}, KSH~\cite{ksh}, ACQ, and CMFH.

For the MIRFLICKR dataset, we choose to compare with hand-crafted feature based algorithms: CVH, CMSSH, CMFH, SePH. We also compare with DCMH, which is an end-to-end approach. DCVH-THN and DCVH-CVH are included in the comparison as well. 2,000 samples are randomly picked as the query set and the rest are treated as database; meanwhile 5,000 randomly selected samples are used for training.

We evaluate the quality of the proposed DCVH via retrieval task and annotation task. Mean average precision (mAP) and precision-recall curve are adopted as evaluation metrics for Hamming ranking and hash lookup retrieval procedures, respectively. The prediction of tags or annotations for images can be obtained directly via the linear classifier $\mathbf{W}^{\mathcal{I}}$ of the proposed algorithm. Based on the $K$ highest ranked prediction, the overall precision (O-P), recall (O-C) and F1-score (O-F1) are used as evaluation metrics:
\begin{align}
\text{O-P} = \frac{N_{CP}}{N_P},\ \text{O-R} = \frac{N_{CP}}{N_G},\ \text{O-F1} = 2\frac{\text{O-P}\cdot \text{O-R}}{\text{O-P}+\text{O-R}}
\end{align}
where $C$ is the number of tags/annotations; $N_{CP}$ is the number of correctly predictions for validation set; $N_P$ is the total number of predictions; $N_G$ is the total number of ground truth for validation set.

\subsection{Results of Cross-view Retrieval Task}

\subsubsection{Cross-view Retrieval}

For DCVH and all the compared baselines, the cross-view retrieval performance based on Hamming ranking on all the datasets is reported in Table~\ref{tab:map_results}, including the task of image retrieval with text query and text retrieval with image query. We can observe that DCVH generally outperforms the compared methods on the two benchmark datasets with various code lengths. For the MS COCO datasets, since DCMH does not provide results, we can observe that on image query retrieving from text database, DCVH outperforms its two variations DCVH-THN and DCVH-CVH by $11\%$ and $0.5\%$, respectively; on text query retrieving from image database, DCVH improves DCVH-THN and DCVH-CVH by $12\%$ and $1\%$ for all code lengths, respectively. DCVH also outperforms the hand-crafted feature based algorithms by at least $16\%$ on both cross-view retrieval tasks. For the MIRFLICKR dataset, DCVH outperforms DCMH on image query retrieving from text database task by $0.2\%$, $2.5\%$, and $3.9\%$ for code length of 16 bits, 32 bits, and 64 bits, respectively; on text query retrieving from image database task, DCVH improves DCMH by $2.2\%$, $4\%$ and $4\%$ for code length of 16 bits, 32 bits, and 64 bits, respectively. It also outperforms hand-crafted feature based algorithms such as SePH by from around $4\%$ to $9\%$ on both cross-view retrieval tasks over various code lengths.

Furthermore, we use hash lookup to compare DCVH with its two variations DCVH-THN and DCVH-CVH to validate the effectiveness of the view-alignment used in DCVH. SePH is also included into the comparison as the baseline. The comparison is summarized in Fig.~\ref{fig:prcurve} in terms of precision-recall curve. For the MS COCO dataset, although DCVH obtains slightly lower precision for higher recall level on two cross-view retrieval tasks, it outperforms its variations and SePH in general. For the MIRFLICKR dataset, DCVH achieves the highest precision at all recall level comparing to other methods.

The comparison of mAP and precision-recall curve on the two benchmark datasets confirms the superiority of the proposed DCVH. As an end-to-end approach, DCVH not only captures the rich semantics from images using CNN, it also is able to extract textual information from pretrained GloVe vectors by using text-CNN. Furthermore, the comparison results of DCVH and its variations validates that the view alignment employed by DCVH is more effective.

\subsubsection{Single-view Image retrieval}\label{sec:singleimage}

As suggested by ACQ~\cite{acq}, cross-view hashing can improve single-view similarity search. This is because the semantic information of the textual data is carries over to image view thanks to the view alignment. We compare DCVH with several state-of-the-art supervised image hashing algorithms, as well as with cross-view image hashing algorithms. Similar to the cross-view image hashing, the comparison result is reported in terms of mAP, as shown in Table~\ref{tab:mscoco_map}. We can observe that DCVH improves HashNet by around $2.5\%$ across various code lengths on MS COCO dataset. Similar to DCVH, DBE also learns binary representation via classification. DCVH outperforms DBE around $5\%$ with different code lengths.

\begin{table}
\centering
\scalebox{1.0}{
\begin{tabular}{ l c c c c c}
	\toprule
	Code length		 		& 16 bits 	&	32 bits	& 48	 bits	&	64 bits	 \\
	\midrule
	CMFH	~\cite{cmfh}		&0.476	&0.484	&0.497	&0.505		\\
	CCA-ACQ~\cite{acq}	&0.500	&0.504	&0.515	&0.520	\\
	KSH~\cite{ksh}			&0.521 &0.534 &0.534 &0.536\\
	DHN~\cite{dhn}			&0.677 	& 0.701 	& 0.695 &0.694\\	
	DBE~\cite{dbe}			&0.623	&0.670	&0.692	&0.716\\
	HashNet~\cite{hashnet}	&\underline{0.687}	&\underline{0.718}	&\underline{0.730}	&\underline{0.736}\\
	DCVH				&{\bf 0.721}		&{\bf 0.748} &{\bf 0.757}	&{\bf 0.761}\\
	\bottomrule
\end{tabular}
}
\caption{Comparison of mean average precision (mAP) on COCO for single-view image retrieval task}
\label{tab:mscoco_map}
\end{table}

\subsection{Image Annotation}

We compare DCVH with several state-of-the-art multilabel image annotation algorithms including WARP~\cite{warp} and DBE~\cite{dbe} on the MS COCO dataset. Note that the performance is evaluated on validation set of MS COCO, which involves 40K samples. Using overall-precision (O-P), overall-recall (O-R) and overall-F1 score (O-F1), the results are based on top-$3$ prediction of annotations, and are summarized in Table~\ref{tab:coco_pred}. From the table we can see that DCVH is able to provide competitive results for image annotation task. This suggests that despite the compromise for the view alignment, DCVH still manages to provide strong performance on image annotation/tagging task. Comparing to its variations, DCVH presents slightly improved performance, suggesting that our proposed view alignment causes minimal interference during the learning of discriminative information. This shows that the binary representation learned by DCVH can be used for different visual tasks, making DCVH a multitasking binary representation learning method.
\begin{table}[!htbp]
\centering
\scalebox{1.0}{
\begin{tabular}{ l | c c c  }
	\toprule
	Method 							& O-P	& 		O-R		&	O-F1   \\
	\hline
	WARP~\cite{warp}		&0.598 	& 0.614			& 	0.606 \\
	DBE~\cite{dbe} (64 bits)			& 0.595	& \textbf{0.627}			&	\textbf{0.611}	\\
	DCVH-THN (64 bits)	& 0.596			& 0.615							& 0.605\\
	DCVH-CVH (64 bits)	& 0.583			& 0.604							& 0.594\\
	DCVH (16 bits)	& 	0.546				& 0.563		& 0.554			\\
	DCVH (32 bits)	& 0.572		&	0.591				& 0.581				\\
	DCVH (64 bits)	& {\bf 0.601}			& \underline{0.617}		& \underline{0.609}			\\
	\bottomrule
\end{tabular}
}
\caption{Performance comparison on MS COCO for image annotation task, compared on top-3 predictions.}
\label{tab:coco_pred}
\end{table}

\subsection{Hyperparameter Selection}\label{sec:hyper}

\begin{figure}
	\centering
    		  \includegraphics[width=3.2in]{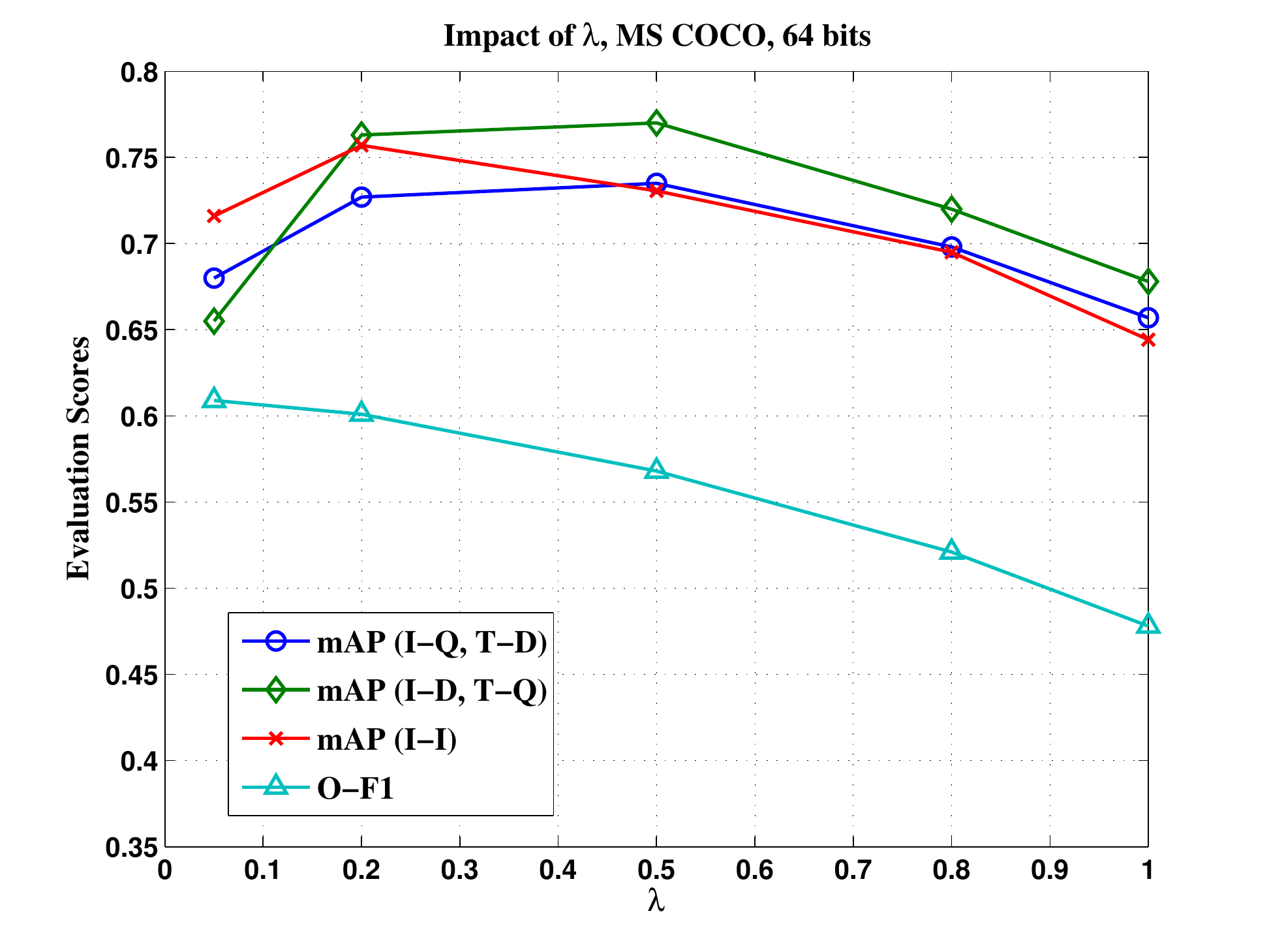}
  	\caption{Impact of $\lambda$ on DCVH, evaluated on several tasks including image query retrieving from text database (I-Q, T-D); text query retrieving from image database (I-D, T-Q); image query retrieving from image database (I-I); and image annotation. mAP is used as evaluation metric for retrieval tasks and overall F1 score (O-F1) is used for annotation task. The code length is set as 64 bits across the experiments.}
\label{fig:lambda}
\end{figure}

In this section, we provide experimental analysis on the hyperparameter $\lambda$. Generally $\lambda$ controls the strength of view alignment and the discriminativeness of the learned binary representation. For the purpose of showing such characteristics, we set $\lambda$ to different value ($0.05,\  0.2,\ 0.5,\ 0.8,\ 1.0$) and evaluate its impact according to the performance of DCVH on cross-view retrieval, single-view image retrieval and image annotation tasks. All the experiments are conducted on MS COCO. Fig.~\ref{fig:lambda} summarizes the results w.r.t. various $\lambda$ values.

It can be observed from Fig.~\ref{fig:lambda} that the cross-view retrieval performance dwindles when $\lambda$ is either too large or too small. This is because when $\lambda \rightarrow 1$, the two views are strongly aligned while the discriminative information (supervision from labels) being very weak. Consequently the semantics from images and views cannot be well preserved in the binary representations. On the contrary, when $\lambda\rightarrow 0$, the semantics of the two views are not aligned well, leading to poor performance on retrieval across different views. Interestingly, the single-view image retrieval performance is enhanced when $\lambda$ is near 0.2. We argue that the semantics from texts provides extra information for similarity search, although the image retrieval performance generally goes down with the increasing $\lambda$. Finally, we can see that image annotation is best performed by DCVH when $\lambda$ is near 0. And its performance goes down more significantly especially when $\lambda$ is greater than 0.2. By considering different tasks performed by DCVH, we choose to set $\lambda=0.2$.

\section{Conclusions}\label{sec:con}
In this paper, we proposed a binary representation learning method: Discriminative Cross-View Hashing. It is an end-to-end approach for effective multimedia content understanding, and includes several novelties to learn high-quality binary representations. First, for both image-view and text-view, it uses CNN-based nonlinear projections, together with Direct Binary Embedding (DBE) layers to learn effective hashing functions, yielding high-quality binary codes; second, it exploits an effective view alignment scheme, which applies bit-wise XOR operation for Hamming distance minimization directly on the continuous embedding during training. Extensive experiments conducted on two benchmark datasets suggest that DCVH learns effective multitasking binary representation that provides superior performance on various computer vision tasks, including cross-view retrieval, (single-view) supervised image retrieval and image annotation.

{\small
\bibliographystyle{ieee}
\bibliography{abits}

\begin{thebibliography}{10}\itemsep=-1pt

\bibitem{45166}
M.~Abadi, A.~Agarwal, P.~Barham, E.~Brevdo, Z.~Chen, C.~Citro, G.~Corrado,
  A.~Davis, J.~Dean, M.~Devin, S.~Ghemawat, I.~Goodfellow, A.~Harp, G.~Irving,
  M.~Isard, Y.~Jia, R.~Jozefowicz, L.~Kaiser, M.~Kudlur, J.~Levenberg,
  D.~Mané, R.~Monga, S.~Moore, D.~Murray, C.~Olah, M.~Schuster, J.~Shlens,
  B.~Steiner, I.~Sutskever, K.~Talwar, P.~Tucker, V.~Vanhoucke, V.~Vasudevan,
  F.~Viégas, O.~Vinyals, P.~Warden, M.~Wattenberg, M.~Wicke, Y.~Yu, and
  X.~Zheng.
\newblock Tensorflow: Large-scale machine learning on heterogeneous distributed
  systems, 2015.

\bibitem{cmssh}
M.~M. Bronstein, A.~M. Bronstein, F.~Michel, and N.~Paragios.
\newblock Data fusion through cross-modality metric learning using
  similarity-sensitive hashing.
\newblock In {\em 2010 IEEE CVPR}, June 2010.

\bibitem{dvsh}
Y.~Cao, M.~Long, J.~Wang, Q.~Yang, and P.~S. Yu.
\newblock Deep visual-semantic hashing for cross-modal retrieval.
\newblock In {\em Proc. KDD}, 2016.

\bibitem{THN}
Z.~Cao, M.~Long, J.~Wang, and Y.~Qiang.
\newblock Transitive hashing network for heterogeneous multimedia retrieval.
\newblock In {\em Proc. AAAI}, 2017.

\bibitem{hashnet}
Z.~{Cao}, M.~{Long}, J.~{Wang}, and P.~S. {Yu}.
\newblock {HashNet: Deep Learning to Hash by Continuation}.
\newblock {\em ArXiv e-prints}, Feb. 2017.

\bibitem{cmfh}
G.~Ding, Y.~Guo, and J.~Zhou.
\newblock Collective matrix factorization hashing for multimodal data.
\newblock In {\em 2014 IEEE CVPR}, pages 2083--2090, June 2014.

\bibitem{warp}
Y.~Gong, Y.~Jia, T.~K. Leung, A.~Toshev, and S.~Ioffe.
\newblock {Deep Convolutional Ranking for Multilabel Image Annotation}.
\newblock In {\em ICLR}, 2014.

\bibitem{Hardoon:2004:CCA:1119696.1119703}
D.~R. Hardoon, S.~R. Szedmak, and J.~R. Shawe-taylor.
\newblock Canonical correlation analysis: An overview with application to
  learning methods.
\newblock {\em Neural Comput.}, 16(12), Dec. 2004.

\bibitem{He_2016_CVPR}
K.~He, X.~Zhang, S.~Ren, and J.~Sun.
\newblock Deep residual learning for image recognition.
\newblock In {\em The IEEE Conference on Computer Vision and Pattern
  Recognition (CVPR)}, June 2016.

\bibitem{huiskes08}
M.~J. Huiskes and M.~S. Lew.
\newblock The mir flickr retrieval evaluation.
\newblock In {\em MIR '08: Proceedings of the 2008 ACM International Conference
  on Multimedia Information Retrieval}, New York, NY, USA, 2008. ACM.

\bibitem{acq}
G.~Irie, H.~Arai, and Y.~Taniguchi.
\newblock Alternating co-quantization for cross-modal hashing.
\newblock In {\em The IEEE International Conference on Computer Vision (ICCV)},
  December 2015.

\bibitem{dcmh}
Q.~Jiang and W.~Li.
\newblock Deep cross-modal hashing.
\newblock {\em CoRR}, abs/1602.02255, 2016.

\bibitem{joulin2016bag}
A.~Joulin, E.~Grave, P.~Bojanowski, and T.~Mikolov.
\newblock Bag of tricks for efficient text classification.
\newblock {\em arXiv preprint arXiv:1607.01759}, 2016.

\bibitem{DBLP:conf/nips/KrizhevskySH12}
A.~Krizhevsky, I.~Sutskever, and G.~E. Hinton.
\newblock Imagenet classification with deep convolutional neural networks.
\newblock In {\em Advances in Neural Information Processing Systems}, pages
  1106--1114, 2012.

\bibitem{cvh}
S.~Kumar and R.~Udupa.
\newblock Learning hash functions for cross-view similarity search.
\newblock In {\em Proc. IJCAI}, pages 1360--1365. AAAI Press, 2011.

\bibitem{Lin2014}
T.-Y. Lin, M.~Maire, S.~Belongie, J.~Hays, P.~Perona, D.~Ramanan,
  P.~Doll{\'a}r, and C.~L. Zitnick.
\newblock {\em Microsoft COCO: Common Objects in Context}, pages 740--755.
\newblock Springer International Publishing, Cham, 2014.

\bibitem{seph}
Z.~Lin, G.~Ding, M.~Hu, and J.~Wang.
\newblock Semantics-preserving hashing for cross-view retrieval.
\newblock In {\em The IEEE Conference on Computer Vision and Pattern
  Recognition (CVPR)}, June 2015.

\bibitem{sentence_lstm}
Z.~Lin, M.~Feng, C.~N. dos Santos, M.~Yu, B.~Xiang, B.~Zhou, and Y.~Bengio.
\newblock A structured self-attentive sentence embedding.
\newblock In {\em ICLR 2017 (Conference Track)}, 2017.

\bibitem{Liu_2016_CVPR}
H.~Liu, R.~Wang, S.~Shan, and X.~Chen.
\newblock Deep supervised hashing for fast image retrieval.
\newblock In {\em The IEEE Conference on Computer Vision and Pattern
  Recognition (CVPR)}, June 2016.

\bibitem{cebits}
L.~Liu and H.~Qi.
\newblock Learning effective binary descriptors via cross entropy.
\newblock In {\em 2017 IEEE Winter Conference on Applications of Computer
  Vision (WACV)}, pages 1--8. IEEE, 2017.

\bibitem{dbe}
L.~Liu, A.~Rahimpour, A.~Taalimi, and H.~Qi.
\newblock {End-to-end Binary Representation Learning via Direct Binary
  Embedding}.
\newblock {\em ArXiv e-prints}, Mar. 2017.

\bibitem{ksh}
W.~Liu, J.~Wang, R.~Ji, Y.-G. Jiang, and S.-F. Chang.
\newblock Supervised hashing with kernels.
\newblock In {\em CVPR}, pages 2074--2081, 2012.

\bibitem{word2vec}
T.~Mikolov, I.~Sutskever, K.~Chen, G.~Corrado, and J.~Dean.
\newblock Distributed representations of words and phrases and their
  compositionality.
\newblock In {\em Proc. NIPS}, NIPS'13, pages 3111--3119, USA, 2013. Curran
  Associates Inc.

\bibitem{pennington2014glove}
J.~Pennington, R.~Socher, and C.~D. Manning.
\newblock Glove: Global vectors for word representation.
\newblock In {\em Empirical Methods in Natural Language Processing (EMNLP)},
  pages 1532--1543, 2014.

\bibitem{Shen_2015_CVPR}
F.~Shen, C.~Shen, W.~Liu, and H.~Tao~Shen.
\newblock Supervised discrete hashing.
\newblock In {\em The IEEE Conference on Computer Vision and Pattern
  Recognition (CVPR)}, June 2015.

\bibitem{Wang_2016_CVPR}
J.~Wang, Y.~Yang, J.~Mao, Z.~Huang, C.~Huang, and W.~Xu.
\newblock Cnn-rnn: A unified framework for multi-label image classification.
\newblock In {\em CVPR}, June 2016.

\bibitem{spectralhashing}
Y.~Weiss, A.~Torralba, and R.~Fergus.
\newblock Spectral hashing.
\newblock In D.~Koller, D.~Schuurmans, Y.~Bengio, and L.~Bottou, editors, {\em
  NIPS}, pages 1753--1760. Curran Associates, Inc., 2009.

\bibitem{text_cnn}
J.~Xu, P.~Wang, G.~Tian, B.~Xu, J.~Zhao, F.~Wang, and H.~Hao.
\newblock Convolutional neural networks for text hashing.
\newblock In {\em Proc. IJCAI}, pages 1369--1375. AAAI Press, 2015.

\bibitem{dhn}
H.~Zhu, M.~Long, J.~Wang, and Y.~Cao.
\newblock Deep hashing network for efficient similarity retrieval.
\newblock In {\em Proc. AAAI}, pages 2415--2421, 2016.

\bibitem{Zhuang_2016_CVPR}
B.~Zhuang, G.~Lin, C.~Shen, and I.~Reid.
\newblock Fast training of triplet-based deep binary embedding networks.
\newblock In {\em The IEEE Conference on Computer Vision and Pattern
  Recognition (CVPR)}, June 2016.

\end{thebibliography}
}

\end{document}